\pdfoutput=1
\documentclass[11pt]{article}
\usepackage[preprint]{acl}

\usepackage{times}
\usepackage{latexsym}

\usepackage[T1]{fontenc}

\usepackage[utf8]{inputenc}

\usepackage{microtype}

\usepackage{inconsolata}

\usepackage{graphicx}

\usepackage[utf8]{inputenc}
\usepackage[T1]{fontenc}
\usepackage{hyperref}
\usepackage{url}
\usepackage{booktabs}
\usepackage{amsfonts}
\usepackage{nicefrac}
\usepackage{microtype}
\usepackage{xcolor}
\usepackage{graphicx}
\usepackage{subcaption}
\usepackage{array}
\usepackage{multirow}
\usepackage{enumitem}
\newcolumntype{L}[1]{>{\raggedright\arraybackslash}m{#1}}
\newcolumntype{C}[1]{>{\centering\arraybackslash}m{#1}}

\usepackage{amsmath}
\usepackage{amssymb}
\usepackage{mathtools}
\usepackage{amsthm}
\usepackage{tabularx}
\usepackage{makecell}
\usepackage{placeins}
\usepackage{soul}
\usepackage{tikz}

\usepackage{algorithm}
\usepackage{algorithmic}

\usepackage[capitalize,noabbrev]{cleveref}
\usepackage[most]{tcolorbox}
\tcbuselibrary{breakable}
\newtcolorbox{casebox}[1]{%
  enhanced, colback=blue!3, colframe=blue!40!black,
  fonttitle=\bfseries, title={#1},
  rounded corners, boxrule=0.8pt, breakable,
  top=6pt, bottom=6pt, left=6pt, right=6pt
}
\newtcolorbox{chainbox}[1]{%
  enhanced, colback=gray!6, colframe=gray!50,
  fonttitle=\bfseries\small, title={#1},
  rounded corners, boxrule=0.5pt,
  top=4pt, bottom=4pt, left=5pt, right=5pt,
  before skip=6pt, after skip=6pt
}
\newtcolorbox{taskrule}{%
  enhanced, breakable,
  colback=white,
  colframe=white,
  boxrule=0pt,
  borderline west={2pt}{0pt}{blue!40!black},
  left=8pt, right=0pt, top=3pt, bottom=3pt,
  before skip=2pt, after skip=8pt,
  fontupper=\small,
}
\usepackage{fvextra}
\usepackage{fontawesome5}
\usepackage{wrapfig}
\theoremstyle{plain}

\theoremstyle{definition}

\theoremstyle{remark}

\usepackage[textsize=tiny]{todonotes}

\newcommand{\methodname}{TreeSeeker}

\title{
\methodname{}: Tree-Structured Trial, Error, and Return in Deep Search
}

\author{
Zhuofan Shi$^{1{,}}$\thanks{Equal contribution.}$^{{,}}$\thanks{This work was completed during the authors' internship at Microsoft.},
Mingzhe Ma$^{1{,}2{,}}$\footnotemark[1]$^{{,}}$\footnotemark[2],
Lu Wang$^{1}$,
Fangkai Yang$^{1}$,
Pu Zhao$^{1}$,
Yiming Guan$^{1{,}}$\footnotemark[2],
\\
\bfseries
Youling Huang$^{1{,}}$\footnotemark[2],
Wei Zhang$^{2}$,
Qingwei Lin$^{1}$,
Dongmei Zhang$^{1}$,
Saravan Rajmohan$^{1}$
\\[1ex]
$^{1}$Microsoft
\quad
$^{2}$East China Normal University
\\
\texttt{wlu@microsoft.com}
\quad
\texttt{zhangwei.thu2011@gmail.com}
}

\begin{document}

\maketitle

\begin{abstract}
Deep search requires agents to answer complex questions through multi-step web search, browsing, evidence comparison, and synthesis.
A central challenge is deciding how to search when several directions look plausible but only some will later lead to reliable evidence.
If an agent greedily follows the current best-looking direction, it may keep extending a weak continuation.
If it explores without discipline, it may waste budget on disconnected trials.
We propose \methodname{}, an inference-time framework for controlled trial-and-error in deep search.
\methodname{} organizes search as branch-and-return search over tree-structured states, where each branch is a tentative direction for a sub-goal.
At each round, \textbf{TreeSearch} reads all sub-goal trees, identifies active goals, and uses textual UCB signals of value, uncertainty, and risk to select among exploiting a promising branch, exploring an uncertain alternative, or pruning an unproductive continuation and returning to an earlier branch point.
\textbf{TreeMem} supports this control loop by keeping evidence, uncertainty, conflicts, progress, and failure cues attached to the branches that produced them, so trial outcomes can guide later decisions.
Experiments on XBench-DeepSearch, BrowseComp, and BrowseComp-ZH show that \methodname{} consistently outperforms strong open-source baselines, suggesting that explicit branch-and-return control complements stronger reasoning and tool execution. 
\end{abstract}

\section{Introduction}

Deep search is a long-horizon information-seeking task where an agent answers complex questions by interacting with web information sources over multiple steps. 
Unlike closed-book question answering or single-step retrieval, deep search requires the agent to plan subgoals~\citep{nie2025parallelresearch,qin2025flashsearcher}, issue search queries, browse webpages~\citep{wu2025webdancer,li2025webthinker}, inspect evidence~\citep{chen2025iterresearch,ye2025agentfold}, resolve conflicts, and synthesize a final answer. 
LLM-based agents are a natural fit for this setting because they can combine reasoning with tool use, as shown by ReAct-style agents~\citep{yao2022react} that interleave reasoning and actions. 
As the search horizon grows, however, the main challenge is not only choosing the next action, but deciding when to try a new direction, continue a useful one, or return from a failed one.

Recent work has improved deep-search agents in several ways. 
Agentic training and post-training~\citep{team2025tongyi,lu2025deepdive,li2025websailor,wu2025webdancer} help models plan, search, browse, and synthesize evidence over long trajectories. 
Context and memory methods reduce the burden of long histories by compressing past interactions~\citep{wu2025resum} or reconstructing an evolving workspace across research rounds~\citep{chen2025iterresearch,ye2025agentfold}. 
Inference-time execution methods improve efficiency by decomposing tasks into structured subgoals and executing independent parts in parallel~\citep{nie2025parallelresearch,qin2025flashsearcher}. 
These methods make agents stronger and more efficient, but they do not fully address how an agent should control search when multiple directions look plausible. 
The agent still needs to decide which direction to continue, which alternative to try, and when to stop a weak continuation and return to an earlier branch point.

This control problem appears because early search directions are often uncertain. 
A webpage, source family, query formulation, or intermediate hypothesis may look useful at first but later lead to weak, conflicting, or incomplete evidence. 
If the agent greedily follows the current best-looking direction, it may keep extending a weak continuation. 
If it tries alternatives without a clear rule, it may waste budget on disconnected attempts. 
Effective deep search therefore needs controlled trial-and-error, where the agent can continue useful directions, test uncertain alternatives, and return from unproductive continuations under a limited budget.

We propose \methodname{}, an inference-time framework that organizes deep search as \emph{branch-and-return search}. 
Each branch represents a search direction, such as a query, source family, or hypothesis, and each tree maintains the branches for a sub-goal. 
At each decision round, \methodname{} reads all sub-goal trees, identifies dependency-ready unresolved goals, and outputs one operation for each active goal in a single pass. 
Its controller, \textbf{TreeSearch}, uses an operation-level textual UCB rule to allocate search budget across exploiting a promising branch, exploring an uncertain alternative, or pruning an unhelpful continuation and returning to an earlier branch point. 
The rule compares three ordinal semantic signals, \emph{value}, \emph{uncertainty}, and \emph{risk}, and extends the explore--exploit principle with a risk-aware return mechanism.
\textbf{TreeMem} supports this control loop by keeping evidence, uncertainty, conflicts, progress, and failure cues attached to the branches that produced them. 
Together, TreeSearch and TreeMem allow the agent to compare tentative directions, continue useful ones, test uncertain alternatives, and recover from unhelpful search paths without repeatedly extending a single linear trajectory.

We evaluate \methodname{} on XBench-DeepSearch~\citep{xbench}, BrowseComp~\citep{browsecomp}, and BrowseComp-ZH~\citep{browsecomp-zh}.
\methodname{} achieves 56.3 on XBench-DS, 47.0 on BrowseComp, and 43.0 on BrowseComp-ZH, achieving the best performance among the evaluated open-source baselines.
Ablations on XBench-DS show that removing textual UCB signals and disabling the branch-and-return operations reduce performance by 4.3 and 8.3 points, respectively, confirming that both operation-level scoring and branch-and-return control are necessary.

In summary, our contributions are:
\begin{itemize}[leftmargin=*]
\item We propose \methodname{}, a branch-and-return framework that organizes deep search over tree-structured states. At each decision round, it reads all sub-goal trees and decides one operation for each active sub-goal.

\item We introduce TreeSearch and TreeMem as the two core components of \methodname{}. TreeSearch performs operation-level textual UCB control using value, uncertainty, and risk signals, while TreeMem keeps evidence, uncertainty, conflicts, progress, and failure cues attached to their corresponding branches.

\item We evaluate \methodname{} on XBench-DeepSearch, BrowseComp, and BrowseComp-ZH, with ablations and cost analysis to study the effects of branch-and-return control, textual UCB-style decision making, and structured branch memory.
\end{itemize}

\section{Related Work}

\subsection{Single-Path deep search Agents and the Premature Commitment Problem}

A dominant paradigm in deep-search agent design is the sequential reason-act-observe loop, exemplified by ReAct~\citep{yao2022react}, where the agent maintains a single evolving trajectory of plans, tool calls, and observations. IterResearch~\citep{chen2025iterresearch}, WebSailor~\citep{li2025websailor}, WebDancer~\citep{wu2025webdancer}, and WebtThinker~\citep{li2025webthinker} strengthen this paradigm through workspace reconstruction, improved web reasoning, or long-horizon training and interaction scaling. Flash-Searcher~\citep{qin2025flashsearcher} and ParallelResearch~\citep{nie2025parallelresearch} further introduce DAG-structured subgoals and parallel execution to improve throughput and coverage. Search More, Think Less~\citep{chen2026searchmorethinkless} takes a complementary direction by replacing deep per-step reasoning with wider evidence acquisition to improve efficiency and generalization. However, these systems still do not provide evidence-driven control for reallocating budget across alternative paths within a goal.

This limitation matters under early-stage uncertainty, a single-path or fixed-schedule agent may continue building on weak evidence rather than redirecting budget based on what has been learned mid-search. \methodname{} differs by treating each candidate path as a persistent semantic branch state and using TreeSearch to decide which branch to deepen, explore, or prune based on accumulated evidence and failure signals.

\subsection{Context Management and Flat History}

Long-horizon agents also face a context bottleneck, motivating summarization and workspace-reconstruction methods. ReSum~\citep{wu2025resum} periodically compresses exploration history, IterResearch~\citep{chen2025iterresearch} reconstructs an evolving workspace across research rounds, MemAgent~\citep{yu2025memagent} rebuilds task state on demand, and AgentFold~\citep{ye2025agentfold} condenses and reorganizes search history to prevent context overload.

These methods address the context bottleneck effectively, but they organize history as a single evolving state. When multiple tentative search directions are mixed into a flat context or a single reconstructed workspace, it becomes difficult to distinguish which attempt was useful, which failed, and which alternative branches remain worth pursuing. \methodname{} uses a different role for summarization: rather than compressing history into one state, TreeMem keeps summarized evidence attached to the branch that produced it. The resulting branch-local states are not memory devices for continued single-path reasoning, but decision objects that TreeSearch can compare, deepen, or prune.

\subsection{Trial-and-Error, UCB, and Tree Search for Language Agents}

The explore--exploit tradeoff formalized by UCB~\citep{auer2002ucb} provides a principled basis for allocating effort among competing alternatives under uncertainty. In the language agent setting, tree-search methods have applied this intuition to reasoning and planning: Language Agent Tree Search (LATS)~\citep{zhou2023lats} unifies reasoning, acting, and planning in an MCTS-style framework, and follow-up works such as Plan-MCTS~\citep{zhang2026planmcts}, ExACT~\citep{yu2024exact}, and uncertainty-aware web-agent frameworks~\citep{zhang2026webuncertainty} apply tree-based exploration to web navigation or interactive decision settings. In retrieval-augmented generation, MCTS-RAG~\citep{tong2025mctsrag} structures retrieval as tree search, Air-RAG~\citep{feng2025airrag} interleaves retrieval with reasoning via expansion and simulation, and ReARTeR~\citep{sun2025rearter} uses MCTS-guided process rewards for preference optimization.

These methods operate over well-defined action spaces such as query reformulations, document selections, or reasoning steps, where branch quality can be estimated with scalar rollout rewards or retrieval scores. Deep search presents a different challenge: the objects being compared are partially synthesized semantic evidence states, not discrete actions, and quality signals include partial answers, source conflicts, unresolved constraints, and failure cues rather than numeric rewards. \methodname{} extends the UCB intuition to this setting through a textual scoring rule that estimates branch value, residual uncertainty, and search risk from semantic branch states, and couples it with a return mechanism that prunes unproductive continuations and redirects search to earlier branch points. This is broadly related to test-time hypothesis exploration in inductive reasoning~\citep{chen2026surveyinductivereasoninglarge}, but our setting differs in that hypotheses are embedded in open-ended web-search branches whose evidence must be actively gathered and revised.

\definecolor{methodblue}{HTML}{2F6FBB}
\definecolor{methodgreen}{HTML}{2E7D32}
\definecolor{methodorange}{HTML}{B45F06}
\definecolor{methodpurple}{HTML}{6F42C1}

\providecommand{\methodbadge}[3]{%
  \tikz[baseline=(char.base)]{%
    \node[shape=circle, draw=#1, fill=#1!10, inner sep=1.15pt] (char)
      {\textcolor{#1}{\scriptsize\bfseries #2}};
  }\,\textcolor{#1}{\textbf{#3}}%
}
\providecommand{\TSrole}{\methodbadge{methodblue}{S}{TreeSearch}}
\providecommand{\TMrole}{\methodbadge{methodgreen}{M}{TreeMem}}
\providecommand{\TUCBrole}{\methodbadge{methodorange}{U}{Textual UCB}}
\providecommand{\RETrole}{\methodbadge{methodpurple}{$\curvearrowleft$}{return}}
\providecommand{\ExploreOp}{\methodbadge{methodorange}{?}{\textsc{Explore}}}
\providecommand{\ExploitOp}{\methodbadge{methodblue}{$\triangleright$}{\textsc{Exploit}}}
\providecommand{\PruneOp}{\methodbadge{methodpurple}{$\times$}{\textsc{Prune}}}

\section{Method}\label{sec:method}

\begin{figure*}[t]
  \centering
  \includegraphics[width=\textwidth]{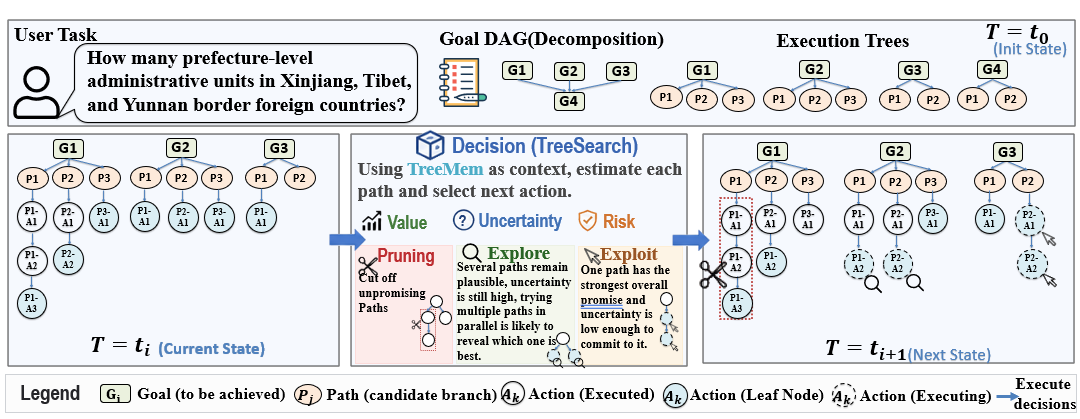}
  \caption{Overview of \methodname{}.
  \textcolor{methodblue}{\textbf{TreeSearch}} is the central trial-and-error controller. It tests uncertain branches, continues promising ones, and prunes weak or misleading attempts.
  \textcolor{methodgreen}{\textbf{TreeMem}} provides the branch-local state required to recognize useful and failed attempts, while the goal DAG determines which goals are eligible for trial-and-error control.}
  \label{fig:method_overview}
\end{figure*}

\subsection{Branch-and-Return Search}
\label{sec:method_overview}

We present \methodname{}, an inference-time framework that organizes deep search as \emph{branch-and-return search}.
As shown in Figure~\ref{fig:method_overview}, given a root question, \methodname{} first decomposes it into sub-goals, since complex tasks often require resolving multiple pieces of evidence before synthesizing the final answer.
Each sub-goal is associated with a search tree.
The first-level branches are candidate paths for solving the sub-goal, such as different queries, sources, or hypotheses, while deeper nodes record the actions and observations produced along each path.
Instead of following one linear search trajectory, \methodname{} keeps these possible directions separated in the trees.
The agent can continue a useful branch, try an uncertain alternative, or prune an unhelpful continuation and return to an earlier branch point for revision.
In this way, trial-and-error is not repeated independent runs, but a structured process over search branches.

\methodname{} has two components.
\TSrole{} decides how the search should move through the trees.
At each decision round, it first reads all sub-goal trees in one shot, identifies dependency-ready unresolved goals, and outputs one operation for each active goal in a single pass.
Here, \emph{operation} denotes the TreeSearch decision (\textsc{Explore}/\textsc{Exploit}/\textsc{Prune}). \emph{Action} denotes the tool action used to execute the selected operation.
\TMrole{} stores compact, failure-aware records for each branch, including evidence, uncertainty, conflicts, progress, and failure cues.
By keeping different search directions separated, these records allow TreeSearch to compare branches, identify useful or failed attempts, and decide whether to explore, exploit, or prune.

Figure~\ref{fig:method_overview} shows the overall loop.
TreeSearch reads the current TreeMem states and outputs one decision for each active sub-goal tree (\textsc{Explore}/\textsc{Exploit}/\textsc{Prune}).
After tool execution or pruning, TreeMem updates the corresponding active branch states.
This loop lets the agent compare search directions, continue useful ones, test alternatives, and recover from unhelpful paths.

Together, TreeMem and TreeSearch turn trial-and-error into branch-and-return search.
We use \TUCBrole{} because trial-and-error needs an explicit control principle. It should not greedily follow the current best-looking branch. It should also not explore alternatives without discipline. A literal path-level UCB formulation would score every path-action pair per sub-goal. That creates a large combinatorial space and turns deep-search control into fine-grained path ranking. We therefore decide at the operation level. For each dependency-ready sub-goal, TreeSearch compares \textsc{Exploit}/\textsc{Explore}/\textsc{Prune} as budget-allocation choices and then grounds the selected operation on concrete path target(s). This preserves the explore and exploit logic, allocates budget across exploit/explore/return in a principled way, and avoids exhaustive path-action enumeration.

\subsection{TreeMem}
\label{sec:treemem}

\paragraph{Tree-structured search state.}
\TMrole{} defines the state interface that TreeSearch reads before each decision round.
For each sub-goal $g_i$, TreeMem stores a tree $\mathcal{T}_i$.
The root stores the \emph{goal state}, including the sub-goal summary and current result candidates.
The first-level nodes are candidate paths for solving the sub-goal, and each path node stores a \emph{branch state}, including evidence, uncertainty, progress, and failure cues.
Deeper nodes store the \emph{recent trace}, including the latest tool calls and returned observations.

\begin{figure}[t]
    \centering
    \includegraphics[width=\columnwidth]{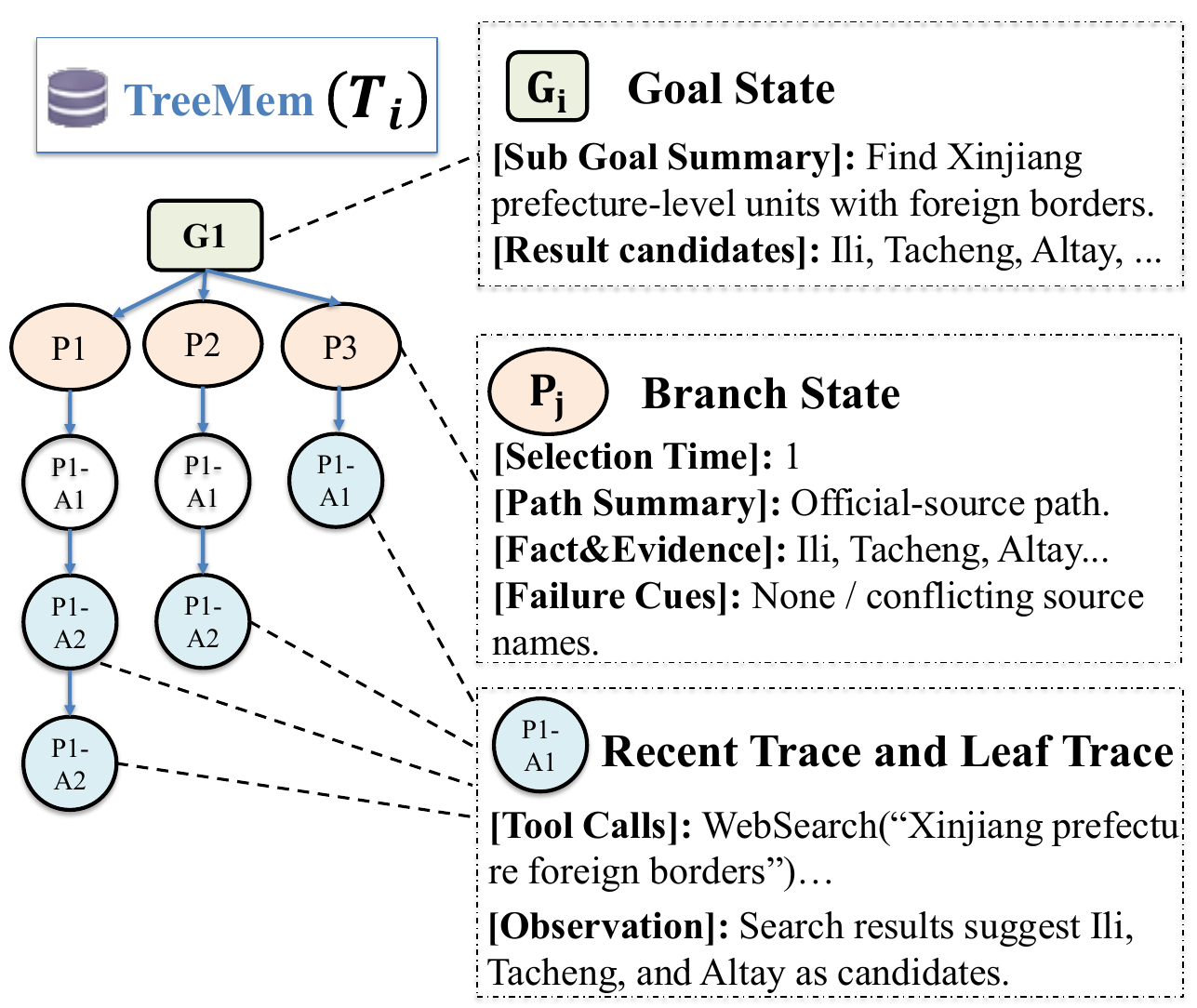}
    \caption{An example of TreeMem.}
    \label{fig:treemem_example}
\end{figure}

Figure~\ref{fig:treemem_example} illustrates this three-level structure.
TreeMem does not keep every trial in full. Long-term history is summarized into goal and branch states. Recent traces keep only the latest raw interactions. Leaf traces store interactions at leaf nodes. Pruned continuations are compressed into short failure cues.
This gives TreeSearch enough structured information to compare branches and decide whether to explore, exploit, or prune without replaying the full interaction history.

\subsection{TreeSearch}
\label{sec:treesearch}

\TSrole{} realizes branch-and-return search as an explicit inference-time control loop.
At each decision round, TreeSearch reads all sub-goal trees, selects dependency-ready unresolved goals, assigns one operation to each selected tree in a single pass, and updates active trees accordingly.
For each selected goal, the decision can be \textsc{Explore}, \textsc{Exploit}, or \textsc{Prune}.
This lets the controller advance multiple sub-goals in parallel while keeping each tree update explicit and local.

At decision round $r$, TreeSearch first reads all sub-goal trees in one shot.
\begin{equation}
  \bar{\mathcal{F}}_r = \{\mathrm{View}(\mathcal{T}_i)\}_{i=1}^{K}.
\end{equation}
It then constructs an actionable frontier over trees that are ready to search but not solved yet.
\begin{equation}
  \begin{aligned}
  \mathcal{F}_r = \{\mathrm{View}(\mathcal{T}_i):\,
  & g_i \text{ is ready to search} \\
  & \text{but not solved yet}\}.
  \end{aligned}
\end{equation}
Here, $\mathrm{View}(\mathcal{T}_i)$ is a compact view of tree $\mathcal{T}_i$, including its active branches and their current states.
Let $\mathcal{S}_r$ be the indices of trees in $\mathcal{F}_r$.
TreeSearch then outputs one decision for every tree in $\mathcal{S}_r$ in a single pass.
Trees outside $\mathcal{S}_r$ are not executed in the current round and remain unchanged.

\paragraph{\textcolor{methodorange}{Textual UCB} operation selection.}
\begin{figure*}[t]
    \centering
    \includegraphics[width=\textwidth]{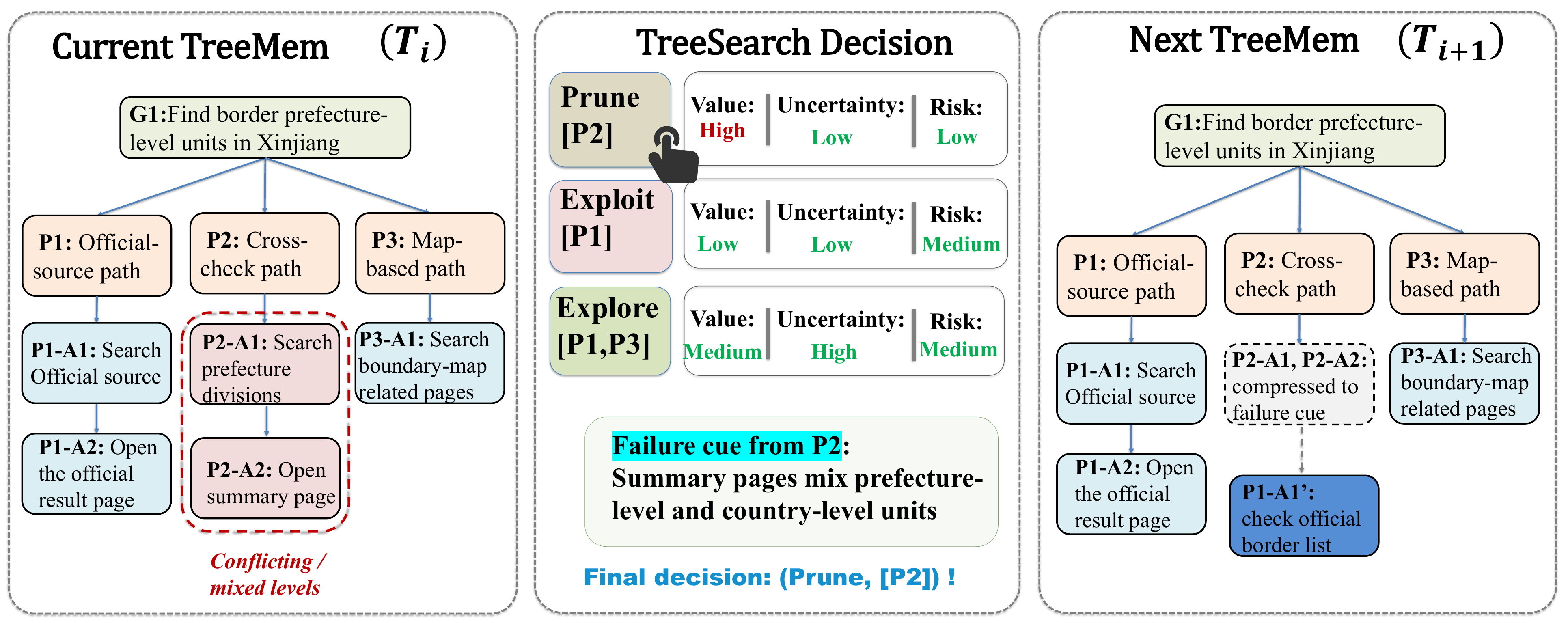}
    \caption{A TreeSearch pruning example. Given the current TreeMem state, TreeSearch uses textual UCB signals to identify a high-risk cross-check continuation, compresses it into a failure cue, and returns to an earlier branch point for revision.}
    \label{fig:treesearch_prune_example}
\end{figure*}

\TUCBrole{} gives TreeSearch a simple rule for controlled trial-and-error.
It avoids both greedy continuation along the current best-looking branch and undisciplined exploration.
It also gives a decision logic for allocating budget across exploit, explore, and return.
For this reason, TreeSearch performs operation selection rather than branch ranking.
For each active goal $g_i$, it compares \textsc{Exploit}, \textsc{Explore}, and \textsc{Prune}, then \emph{binds} each operation to concrete target(s) inside the tree.
This operation binding compresses the action space from unconstrained path-action combinations to compact operation-conditioned candidate sets.
Concretely, we apply two stage filtering before textual scoring. The first stage keeps DAG-ready unresolved goals. The second stage truncates operation-conditioned targets at the path level. This yields a bounded candidate budget $\sum_i\sum_a |\beta_{i,a}|$ with $|\beta_{i,a}| \ll M_i$.
Because branch quality is semantically expressed (partial evidence, source reliability, unresolved conflicts, progress, and failure cues), we use textual rather than numeric UCB.

TreeSearch then applies a textual UCB-style rule defined at the operation level over TreeMem states.
For each candidate operation $a \in \{\textsc{Exploit}, \textsc{Explore}, \textsc{Prune}\}$ with binding $\beta_a$ and induced state $s_a$, it estimates three ordinal semantic signals.
\begin{align}
  \phi(a, \beta_a, s_a) &= (V_a, U_a, R_a), \notag \\
  V_a, U_a, R_a &\in \{\textsc{Low}, \textsc{Medium}, \textsc{High}\}.
\end{align}
Here, $V_a$ (Value) measures expected progress, $U_a$ (Uncertainty) measures expected information gain, and $R_a$ (Risk) measures the chance of committing budget to a misleading continuation.

To make the decision actionable across rounds, we map these ordinal signals to discrete levels $\{0,1,2\}$ (\textsc{Low}$\rightarrow 0$, \textsc{Medium}$\rightarrow 1$, \textsc{High}$\rightarrow 2$), and use a parameter-free score.
\begin{equation}
  \psi(a)=\hat{V}_a + \hat{U}_a - \hat{R}_a,
\end{equation}
where $\hat{V}_a,\hat{U}_a,\hat{R}_a \in \{0,1,2\}$.
TreeSearch selects the operation with the largest $\psi(a)$.
In this way, textual UCB preserves the explore-exploit principle while adapting it to semantically rich branch states and adding a risk-aware return mechanism for branch-and-return search.

Figure~\ref{fig:treesearch_prune_example} illustrates this decision process with a concrete pruning-and-revision case.
For the Xinjiang sub-goal, TreeSearch identifies the cross-check continuation as high-risk because it mixes prefecture-level and county-level units, prunes it into a compact failure cue, and returns to the earlier branch point for revision.

\paragraph{Trial-and-error operations.}
TreeSearch outputs one operation for each active sub-goal tree in the current round.
Together, these operations implement branch-and-return by either expanding a search direction, continuing it, or returning from it.
\begin{equation}
  \begin{aligned}
  \mathcal{D}_r =\,
  &\{(\mathcal{T}_i, b_i, d_i)\}_{i \in \mathcal{S}_r}, \\
  &d_i \in \{\textsc{Explore},\textsc{Exploit},\textsc{Prune}\}.
  \end{aligned}
\end{equation}
Here, $\mathcal{S}_r$ is the active set of dependency-ready unresolved goals, $b_i$ is the selected target in tree $\mathcal{T}_i$, and $d_i$ is the selected operation.
For \textsc{Explore} and \textsc{Exploit}, $b_i$ is the branch to extend.
For \textsc{Prune}, $b_i$ includes the weak continuation to stop and the earlier branch point to return to.

\begin{itemize}[leftmargin=*]
  \item \ExploitOp{} continues a promising branch by extending its current path with a new action.
    For example, if an official-source branch has found a reliable government page, TreeSearch may continue it by extracting the candidate prefecture list.

  \item \ExploreOp{} tests an uncertain alternative by opening a new continuation or sibling branch from the selected branch point.
    For example, if a map-based branch may reveal missing border units but is not yet verified, TreeSearch may open a new continuation to inspect administrative border maps.

  \item \PruneOp{} stops an unhelpful continuation and uses \RETrole{} to move back to an earlier branch point.
    TreeMem keeps a compact failure cue at the pruned continuation and keeps the return point available for later revision.
    For example, if a cross-check continuation repeatedly mixes prefecture-level and county-level units, TreeSearch prunes it and returns to the earlier branch point instead of extending the same weak path.
\end{itemize}

\subsection{TreeSearch and TreeMem Feedback Loop}
\label{sec:feedback_loop}

After TreeSearch outputs the decision set
$\mathcal{D}_r = \{(\mathcal{T}_i, b_i, d_i)\}_{i \in \mathcal{S}_r}$,
each active tree ($i \in \mathcal{S}_r$) is updated independently in the current round.
For \textsc{Explore} and \textsc{Exploit}, the selected branch is passed to the action generator, which executes branch-tagged tool actions.
The resulting observation is written back to the corresponding branch in TreeMem.
For \textsc{Prune}, TreeMem records a failure cue and preserves the earlier branch point for future revision.

The update for each selected tree is written as
\begin{equation}
  \mathcal{T}_i^{r+1}
  =
  \mathrm{Update}(\mathcal{T}_i^r, b_i, d_i, o_i),
  \qquad i \in \mathcal{S}_r,
\end{equation}
where $o_i$ is either the returned observation or the pruning signal.
Trees outside $\mathcal{S}_r$ remain unchanged.
The updated collection of trees is then used in the next decision round.

Together, these operations implement structured trial-and-error, where \textsc{Explore} creates trials, \textsc{Exploit} deepens useful trials, and \textsc{Prune} returns from failed ones.

The main text abstracts this process as a TreeSearch--TreeMem feedback loop; the concrete inference procedure and prompt templates are provided in \Cref{app:treesearch_treemem_loop,app:prompts}.

\section{Experiments}\label{sec:experiments}

\subsection{Experimental Setup}\label{sec:setup}

\paragraph{Benchmarks.}
We evaluate \methodname{} on three public deep-search benchmarks: XBench-DeepSearch~\citep{xbench}, BrowseComp~\citep{browsecomp}, and BrowseComp-ZH~\citep{browsecomp-zh}.
Together, they cover long-horizon web search, compositional browsing, and Chinese-language browsing scenarios.
Due to resource constraints, we randomly sample $100$ instances from BrowseComp and $100$ instances from BrowseComp-ZH as test subsets and report results on these sampled subsets.
Additional benchmark details are provided in \Cref{app:experimental_details}.

\paragraph{Baselines.}
We report results for representative open-source deep-search systems, including Tongyi DeepSearch~\citep{team2025tongyi}, IterResearch~\citep{chen2025iterresearch}, Flash-Searcher~\citep{qin2025flashsearcher}, and LATS~\citep{zhou2023lats}, as well as reported proprietary systems such as OpenAI DeepResearch~\citep{openai_introducing_deep_research_2025} and Gemini DeepResearch~\citep{google_gemini_deep_research}, among other systems listed in Table~\ref{tab:main_result}.

\paragraph{Implementation Details.} 
In this paper, \texttt{gpt-5.2} refers to \texttt{gpt-5.2-20251211}, and \texttt{gpt-4.1} refers to \texttt{gpt-4.1-20250414}. We use \texttt{gpt-5.2} as the default backbone, and additionally evaluate both Flash-Searcher and \methodname{} with \texttt{gpt-4.1} under the same search and browsing tools.
Web search uses the \texttt{Bing Search API v7}~\citep{microsoft_bing_apis}, and page access/parsing uses \texttt{Firecrawl}~\citep{firecrawl2025}; further implementation details are provided in \Cref{app:experimental_details}.

\subsection{Main Results}

\begin{table}[!t]
  \caption{\label{tab:main_result}
    Performance comparison on XBench-DS, BrowseComp, and BrowseComp-ZH. Higher is better.
    For reimplemented baselines and our method, we report the average performance over three independent runs.
    DS = DeepSearch, XB = XBench, BC = BrowseComp.
    $^{*}$ represents reported performance from existing studies.
    }
  \centering
  \footnotesize
  \resizebox{\columnwidth}{!}{%
  \begin{tabular}{@{}l c c c@{}}
    \toprule
    \textbf{Methods} &
    \textbf{XB-DS} &
    \textbf{BC} &
    \textbf{BC-ZH} \\
    \midrule
    \multicolumn{4}{c}{\textbf{Closed-source deep search Systems/Models}} \\
    Gemini-DR*            & 53 & 37.8 & --   \\
    Perplexity Deep Research* & -- & 22.0 & 22.6 \\
    Claude-4-Sonnet-Thinking* & 53.0 & 14.7 & 30.8 \\
    Claude-4.5-Sonnet*   & 66.0 & 19.6 & 40.8 \\
    Gemini-2.5-Pro*      & 56.0 & 9.9  & 32.2 \\
    OpenAI GPT-5*       & 30.0 & 19.8 & 34.3 \\
    OpenAI o1*          & --   & 9.9  & 29.1 \\
    OpenAI o3*          & 68.0 & 55.0 & 59.0 \\
    OpenAI DeepResearch* & 66.7 & 51.5 & 42.9 \\
    Grok3 DeepResearch* & 50+ & 12.9 & --   \\
    Doubao DeepResearch* & 50+ & --   & 26.0 \\
    \midrule
    \multicolumn{4}{c}{\textbf{Open-source deep search Systems/Models}} \\
    DeepSeek-R1*       & 32.7 & 2.0  & 23.2 \\
    Qwen3-235B-A22B-Instruct-2507*        & 45.5 & 8.0  & 23.0 \\
    Kimi K2*                & 54.0 & 11.0 & 22.0 \\
    Search-o1-32B* & 25.0 & 2.8  & 17.9 \\
    WebDancer-32B* & 39.0 & 3.8  & 18.0 \\
    WebSailor-32B* & 53.3 & 10.5 & 25.5 \\
    LATS (gpt-5.2) & 31.0 & 16.0  & 25.7 \\
    Tongyi-DeepSearch-30B-A3B    & 45.0  & 33.3 & 33.0 \\
    IterResearch (gpt-5.2)         & 44.0  & 35.3 & 34.0 \\
    Flash-Searcher (gpt-4.1) & 21.3 & 5.7 & 17.7 \\
    Flash-Searcher (gpt-5.2) & 50.7 & 43.0 & 40.3 \\

    \midrule
    \multicolumn{4}{c}{\textbf{Ours}} \\
    \methodname{} (gpt-4.1)    & 23.0 & 7.7 & 20.3 \\
    \methodname{} (gpt-5.2)    & 56.3 & 47.0 & 43.0 \\
    \bottomrule
  \end{tabular}}
\end{table}

Using \texttt{gpt-5.2}, \methodname{} achieves 56.3 on XBench-DS, outperforming Flash-Searcher, IterResearch, and Tongyi-DeepSearch by 5.6, 12.3, and 11.3 points, respectively.
On BrowseComp and BrowseComp-ZH, it reaches 47.0 and 43.0, ranking first among the evaluated open-source baselines.
Under a shared \texttt{gpt-4.1} backend, \methodname{} also remains ahead of Flash-Searcher, with gains of 1.7, 2.0, and 2.6 points across the three benchmarks, suggesting that the advantage of TreeSearch--TreeMem is not tied to a single backend model.
The strongest controlled comparison is with Flash-Searcher, since both systems use tree-structured search but differ in how candidate paths are controlled after intermediate evidence appears.
The consistent gains over Flash-Searcher under both backbones suggest that explicitly maintaining branch states and selecting among continuation, exploration, and pruning decisions improves the effectiveness of the search process itself.
Taken together, these results show that \methodname{} generalizes well across different deep-search benchmarks, including both English and Chinese settings, and achieves the best performance among the evaluated open-source baselines.
Beyond task accuracy, we also analyze the token usage and tool-call cost of each system in \Cref{sec:cost_analysis}.

\subsection{Cumulative Success over Action Steps}\label{sec:info_coverage}

\begin{figure}[!t]
  \centering
  \includegraphics[width=\linewidth]{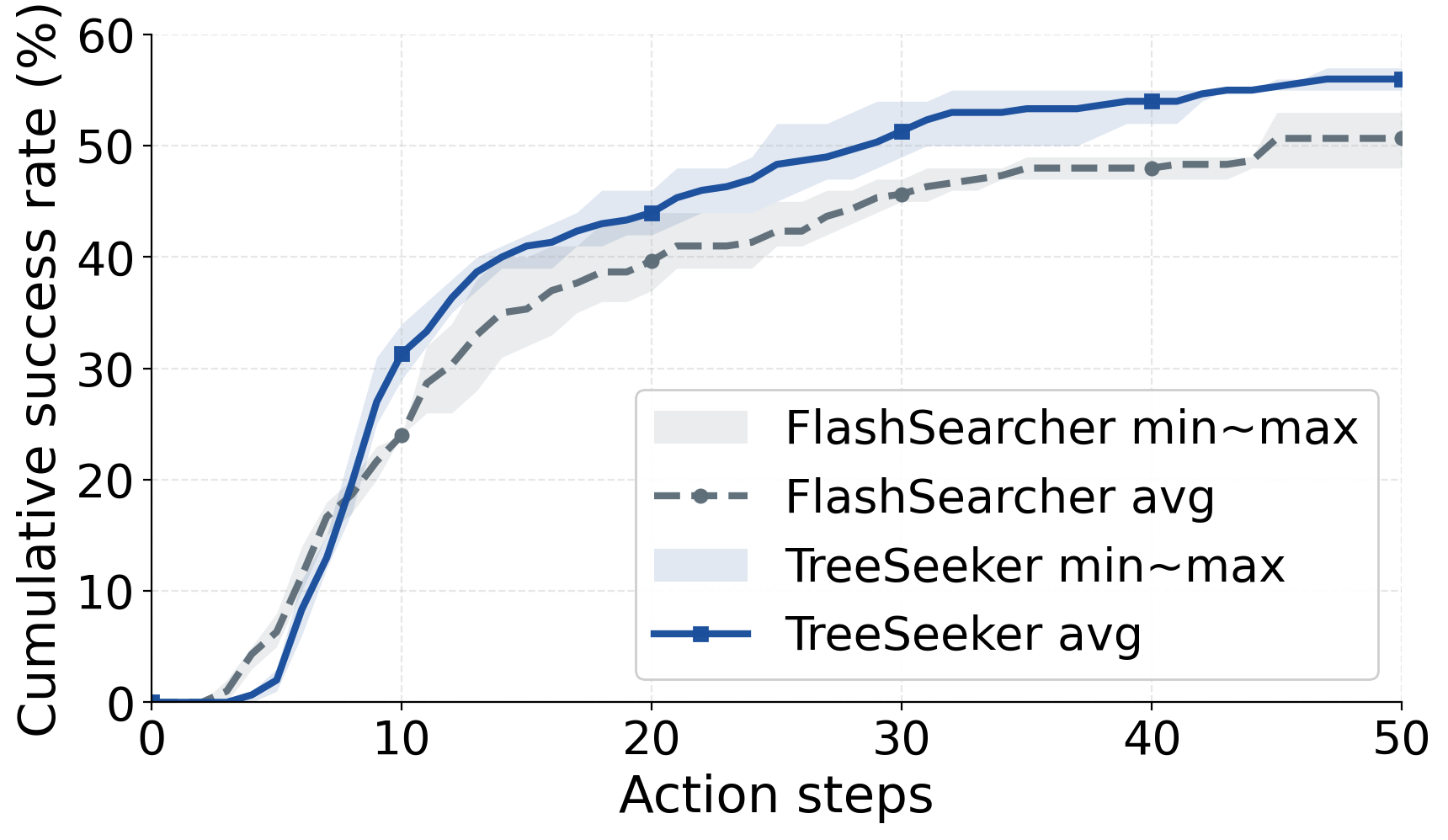}
  \caption{\label{fig:info_coverage}
    Cumulative success rate (\%) versus action steps on XBench-DS.
  }
  \label{fig:info_coverage_success}
\end{figure}

Figure~\ref{fig:info_coverage} shows cumulative success rate versus action steps on the XBench-DS queries shared by all runs, reported as the per-step mean over three runs with min--max envelopes.

The two methods behave similarly in the first several action steps, but their trajectories diverge around step $8$, after which \methodname{} remains consistently ahead.
The gap is visible in both the main growth region (roughly action steps $10$--$30$) and near the end of the budget, where \methodname{} reaches a higher final cumulative success rate.
We attribute this gap to evidence-driven branch selection rather than fixed-schedule path execution. Flash-Searcher executes candidate paths within a goal under a fixed DAG schedule, so later observations have limited effect on whether a path should be continued, revised, or abandoned. In contrast, \methodname{} keeps branch-local evidence, uncertainty, conflicts, and failure cues in TreeMem, and TreeSearch compares \textsc{Exploit}, \textsc{Explore}, and \textsc{Prune} directly over these branch states using textual UCB signals.

\subsection{Ablation Study}

To isolate the contribution of each core component, we conduct ablation experiments on the XBench-DS benchmark.
We consider three ablated variants:
(1)~\textbf{w/o Textual UCB}, which removes the operation-level value, uncertainty, and risk signals from the TreeSearch decision prompt, so the controller selects operations without the textual UCB-style comparison defined in the method section;
(2)~\textbf{w/o Explore \& Prune}, which disables the \textsc{Explore} and \textsc{Prune} operations, so TreeSearch retains only \textsc{Exploit}: it can continue existing branches, but cannot open new continuations or sibling branches, and cannot stop high-risk continuations and return to earlier branch points;
(3)~\textbf{w/o Leaf Trace in TreeMem}, which removes the retained latest raw leaf trace from TreeMem, forcing the agent to rely only on summarized goal and branch states and losing the short-term continuation anchor for each path.

\begin{table}[!t]
  \caption{\label{tab:ablation}
    Ablation study on XBench-DS. Higher is better.
    We report the average performance over three independent runs. Removing Textual UCB scoring, Explore \& Prune operations, or the TreeMem leaf trace leads to consistent performance degradation.
  }
  \centering
  \footnotesize
  \begin{tabular}{l c}
    \toprule
    \textbf{Variant} &
    \textit{avg} \\
    \midrule
    \methodname{} (full) & 56.3 \\
    \methodname{} w/o Textual UCB & 52.0 \\
    \methodname{} w/o Explore \& Prune & 48.0 \\
    \methodname{} w/o Leaf Trace in TreeMem & 51.3 \\
    \bottomrule
  \end{tabular}
\end{table}

Table~\ref{tab:ablation} reports the results. Removing any component degrades performance, showing that textual UCB scoring, branch-and-return operations, and the TreeMem design make complementary contributions.
Removing textual UCB lowers XBench-DS performance from 56.3 to 52.0 ($-$4.3), disabling \textsc{Explore} and \textsc{Prune} causes the largest drop to 48.0 ($-$8.3), and removing the retained leaf trace in TreeMem lowers performance to 51.3 ($-$5.0).
These results indicate that semantic operation scoring, branch-and-return control, and structured branch memory make complementary contributions to effective long-horizon search.

\subsection{Operation Decision Analysis}\label{sec:op_analysis}

Table~\ref{tab:op_freq} reports the empirical frequencies of branch operations on XBench-DS for both the full controller and the \textbf{w/o Textual UCB} ablation.

\begin{table}[t]
  \caption{\label{tab:op_freq}
    Empirical frequencies of branch operations on XBench-DS.
  }
  \centering
  \footnotesize
  \begin{tabular}{@{}l c c c@{}}
    \toprule
     & \textsc{Exploit} & \textsc{Explore} & \textsc{Prune} \\
    \midrule
    \methodname{} & $51.39\%$ & $43.45\%$ & $5.17\%$ \\
    w/o Textual UCB & $36.93\%$ & $61.08\%$ & $1.98\%$ \\
    \bottomrule
  \end{tabular}
\end{table}

With textual UCB guidance, \methodname{} keeps \textsc{Exploit} and \textsc{Explore} relatively balanced ($51.39\%$ vs. $43.45\%$), while using \textsc{Prune} sparingly ($5.17\%$).
Removing textual UCB shifts the controller toward substantially more \textsc{Explore} decisions ($61.08\%$) and fewer \textsc{Exploit} and \textsc{Prune} decisions, suggesting that the semantic control signal helps allocate budget more effectively across deepening, branching, and correction.
Detailed discussion of these tradeoffs is provided in \Cref{app:op_analysis_details}.

\FloatBarrier

\section{Conclusion}

We presented \methodname{}, an inference-time framework that structures deep search as branch-and-return search over tree-structured states.
The key insight is that early-stage decision uncertainty in deep search is best addressed not by a stronger single-path agent, but by maintaining multiple tentative search directions as explicit decision objects and repeatedly allocating budget across them.
\methodname{} realizes this through two tightly coupled components: TreeSearch, which applies an operation-level textual UCB-style rule to decide whether to exploit a promising branch, explore an uncertain alternative, or prune an unproductive continuation and return to an earlier branch point; and TreeMem, which keeps evidence, uncertainty, conflicts, progress, and failure cues attached to the branches that produced them.
Experiments on XBench-DeepSearch, BrowseComp, and BrowseComp-ZH show that \methodname{} consistently outperforms strong open-source baselines, suggesting that explicit branch-and-return control is a valuable complement to stronger reasoning and tool execution in long-horizon deep search.

\section*{Limitations}\label{sec:limitations}

\methodname{} has several limitations.
First, our evaluation is limited to text-based deep search benchmarks.
We do not consider multimodal deep research settings because \methodname{} currently does not integrate multimodal tools such as image or video understanding.
Extending the framework to multimodal evidence sources is left for future work.

Second, branch-and-return control introduces additional inference cost.
TreeSearch requires controller decisions based on value, uncertainty, and risk signals, and TreeMem periodically summarizes branch-local states.
Although our cost analysis shows that \methodname{} remains below Flash-Searcher in total tokens and tool calls, the additional controller and memory updates may still be a concern in latency or budget-sensitive deployments.

Third, \methodname{} relies on external web search and browsing results, which can be noisy, incomplete, outdated, or biased.
TreeSearch and TreeMem help compare evidence, preserve failure cues, and prune unproductive continuations, but they do not guarantee that all retrieved sources are reliable or that all conflicts are fully resolved.
In high-stakes settings, outputs should therefore be checked against trusted sources or human expert review.

\bibliography{custom}

\clearpage

\appendix

\section{\methodname{} Inference Procedure and Implementation Details}
\label{app:treesearch_treemem_loop}

The main text describes \methodname{} as a TreeSearch--TreeMem loop over dependency-ready goal trees.
TreeSearch constructs an active frontier of dependency-ready unresolved goal trees, selects operation-level decisions according to the textual UCB rule, and binds each selected operation to concrete branch target(s).
TreeMem stores the summarized goal and branch states used by later decisions, while a short-term overlay keeps recent unsummarized records and retained leaf traces.
\Cref{alg:treesearch_treemem_loop} specifies the concrete execution procedure used in our implementation, including planning initialization, frontier construction, operation-level decision making, leaf-trace retention, periodic summarization, and final answer generation.

\begin{algorithm}[!t]
\caption{TreeSearch--TreeMem Inference Procedure in \methodname{}}
\label{alg:treesearch_treemem_loop}
\footnotesize
\begin{algorithmic}[1]
\STATE \textbf{Input:} user query $q_0$, decision-round budget $T$, summary interval $H=8$
\STATE \textbf{Output:} final answer $a$
\STATE \textbf{Definitions:} $\mathcal{G}$: sub-goals; $\mathcal{D}_G$: dependency DAG; $\mathcal{P}$: candidate paths
\STATE $\mathcal{M}$: summarized TreeMem; $\mathcal{R}$: recent/leaf-trace overlay; $\Omega$: operation decisions
\STATE $(\mathcal{G}, \mathcal{D}_G, \mathcal{P}) \leftarrow \textsc{PlanAgent}(q_0)$
\STATE $\mathcal{M} \leftarrow \textsc{InitTreeMem}(q_0, \mathcal{G}, \mathcal{D}_G, \mathcal{P})$
\STATE $\mathcal{R} \leftarrow \emptyset$
\FOR{$t = 0$ {\bfseries to} $T-1$}
  \IF{$\textsc{AnswerFound}(\mathcal{M}, \mathcal{R})$}
    \STATE \textbf{break}
  \ENDIF
  \STATE $\mathcal{F}_t \leftarrow \textsc{Frontier}(\mathcal{M},\mathcal{R},\mathcal{D}_G)$
  \STATE $\Omega_t \leftarrow \textsc{TextualUCB}(\mathcal{F}_t,\mathcal{M},\mathcal{R},T-t)$
  \FORALL{$(\mathcal{T}_i,b_i,d_i)\in\Omega_t$}
    \IF{$d_i \in \{\textsc{Explore},\textsc{Exploit}\}$}
      \STATE $(a_i,o_i) \leftarrow \textsc{ExecuteAction}(\mathcal{T}_i,b_i,d_i)$
      \STATE $\mathcal{R} \leftarrow \textsc{UpdateLeafTrace}(\mathcal{R},i,b_i,d_i,a_i,o_i)$
    \ELSE
      \STATE $o_i \leftarrow \textsc{PruneReturn}(\mathcal{T}_i,b_i)$
      \STATE $\mathcal{R} \leftarrow \textsc{RecordPrune}(\mathcal{R},i,b_i,o_i)$
    \ENDIF
  \ENDFOR
  \IF{$(t+1) \bmod H = 0$}
    \STATE $(\mathcal{M},\mathcal{R}) \leftarrow \textsc{Summarize}(\mathcal{M},\mathcal{R})$
  \ENDIF
\ENDFOR
\IF{$\mathcal{R} \neq \emptyset$}
  \STATE $(\mathcal{M},\mathcal{R}) \leftarrow \textsc{Summarize}(\mathcal{M},\mathcal{R})$
\ENDIF
\STATE $a \leftarrow \textsc{Synthesize}(q_0,\mathcal{M})$
\STATE \textbf{return} $a$
\end{algorithmic}
\end{algorithm}

\paragraph{Planning and TreeMem initialization.}
The inference procedure starts with a planning agent.
Given the root query $q_0$, the planner decomposes the task into sub-goals $\mathcal{G}$, constructs a dependency DAG $\mathcal{D}_G$, and proposes candidate paths $\mathcal{P}_i$ for each goal $g_i$.
This follows the formulation in the main text: the DAG controls which goals are dependency-ready, while candidate paths define alternative search directions under each goal.
TreeMem initializes a goal-local tree for each goal, with candidate paths represented as major branches.
In the implementation, $\mathcal{M}$ denotes the summarized TreeMem states, while $\mathcal{R}$ stores recent unsummarized records and retained leaf traces as a short-term overlay.

\paragraph{Frontier construction and textual UCB decision.}
At each iteration, TreeSearch constructs an active frontier from the summarized TreeMem states, the recent-record overlay, and the dependency DAG.
A goal tree is eligible only when its dependencies are satisfied and the goal has not yet been resolved.
TreeSearch then applies the operation-level textual UCB rule described in the main text: for each active goal, it first binds each candidate operation (\textsc{Exploit}, \textsc{Explore}, or \textsc{Prune}) to concrete branch target(s), then scores the bound operations using value, uncertainty, and risk signals.
The resulting decision set contains one bound operation-level decision for each selected active goal tree.
The recent records and retained leaf traces provide short-term context, so the controller can use fresh evidence before it is fully compressed into branch summaries.

\paragraph{Decision execution and leaf-trace update.}
For \textsc{Explore} and \textsc{Exploit}, the execution agent carries out the selected branch action using the available execution tools, including web search, page crawling, and a Python sandbox for code execution.
The resulting action-observation pair is appended to the recent records of the selected path and stored as the latest leaf trace at that path's leaf node, replacing the previous continuation anchor.
For \textsc{Prune}, no external browsing action is required.
TreeMem records a compact failure cue for the weak continuation and preserves the earlier branch point as the return target.
Thus, each path keeps a compact summarized state for long-term decision making and a latest leaf trace or return point for short-term continuation.

\paragraph{Periodic summary updates.}
TreeMem is summarized periodically rather than fully rewritten after every action.
In our implementation, the summary interval is $H=8$ decision rounds.
At each summary step, the summary agent folds recent unsummarized operation records, including action-observation pairs and pruning cues, into the corresponding goal and path states.
The branch summaries are updated with revised evidence, provenance, candidate answers, unresolved constraints, progress signals, conflicts, and failure cues.
After consolidation, older unsummarized records may be compressed or removed to prevent TreeMem from degenerating into a flat growing history.
However, the latest leaf trace of each active path is retained.
This leaf trace is included in the summary update, but it remains explicitly available as the continuation anchor for future TreeSearch decisions and execution.

\paragraph{Termination and answer generation.}
The loop terminates when \textsc{AnswerFound} determines that sufficient evidence has been collected, or when the global step budget is exhausted.
If unsummarized records remain at termination, they are summarized once more before final synthesis.
The final answer is then generated from the updated TreeMem state, including surviving candidate answers, supporting evidence, confidence signals, and unresolved conflict markers when applicable.

\FloatBarrier

\section{Experimental Details}
\label{app:experimental_details}

\paragraph{Benchmarks.}
We evaluate \methodname{} on three publicly available benchmarks that require long-horizon web search, multi-source evidence integration, and complex information synthesis:

\begin{itemize}[leftmargin=*]
  \item \textbf{XBench-DeepSearch}~\citep{xbench} is a benchmark designed to evaluate deep-search capabilities in realistic information-seeking scenarios.
  Its tasks require systems to decompose underspecified questions, perform multiple rounds of search refinement, integrate heterogeneous evidence from different web sources, and synthesize a final answer.

  \item \textbf{BrowseComp}~\citep{browsecomp} is a browsing-based compositional search benchmark released by OpenAI.
  It focuses on hard-to-find, entangled information that cannot usually be answered by a single retrieval step.
  Solving these questions requires formulating effective search queries, navigating search results, inspecting web pages, extracting relevant facts, and synthesizing evidence across multiple sources.
  Due to resource constraints, we randomly sample $100$ BrowseComp instances as a test subset and report results on this sampled subset.

  \item \textbf{BrowseComp-ZH}~\citep{browsecomp-zh} extends the BrowseComp setting to Chinese web environments and information sources.
  It evaluates whether a deep-search system can handle non-English browsing, cross-page reasoning, and culturally or linguistically specific evidence aggregation.
  We include this benchmark to test cross-lingual and cross-cultural generalization beyond English web search.
  Due to resource constraints, we randomly sample $100$ BrowseComp-ZH instances as a test subset and report results on this sampled subset.
\end{itemize}

Together, these benchmarks provide a broad evaluation landscape for testing whether \methodname{} can allocate search budget effectively across English and Chinese browsing tasks, short- and long-horizon evidence gathering, and both single-benchmark and cross-benchmark generalization settings.

\paragraph{Baselines.}
We compare \methodname{} with representative systems that cover both open-source research agents and reported proprietary deep-search products.
The main goal of these comparisons is to distinguish gains from the proposed branch-and-return controller from gains that may come from stronger backend models or larger search budgets.

\begin{itemize}[leftmargin=*]
  \item \textbf{Open-source deep-search systems.}
  Tongyi DeepSearch~\citep{team2025tongyi} is an end-to-end deep-search agent based on its own post-trained Tongyi-A30B-A3B model variant.
  IterResearch~\citep{chen2025iterresearch} improves long-horizon search through iterative workspace reconstruction and interaction scaling.
  Flash-Searcher~\citep{qin2025flashsearcher} uses a DAG-style parallel search framework and is the most direct structural baseline for evaluating whether explicit branch-level control improves over fixed-schedule parallel search.
  LATS~\citep{zhou2023lats} integrates reasoning, acting, and planning in an MCTS-style framework, providing a tree-search baseline for agentic reasoning.

  \item \textbf{Reported open-source model systems.}
  We also include reported results for open-source model or agent systems such as DeepSeek-R1~\citep{deepseek-r1}, Qwen3~\citep{qwen3}, K2~\citep{team2025kimi}, Search-o1~\citep{li2025search}, WebDancer~\citep{wu2025webdancer}, and WebSailor~\citep{li2025websailor} when benchmark numbers are available.
  These results help contextualize the performance of \methodname{} against broader open-source progress in reasoning and web-agent systems.

  \item \textbf{Proprietary deep-search systems and models.}
  We report available results from commercial systems and closed-source models, including OpenAI DeepResearch~\citep{openai_introducing_deep_research_2025}, Gemini DeepResearch~\citep{google_gemini_deep_research}, and other systems listed in Table~\ref{tab:main_result}.
  These comparisons are not controlled for model scale or product-specific infrastructure, but they provide useful reference points for understanding how far an open research framework can approach strong proprietary deep-search systems.
\end{itemize}

\paragraph{Implementation Details.}
For reimplemented open-source systems, we keep the backend model and tool interface aligned whenever the original system design permits, so that differences are mainly attributable to the search-control framework rather than to different web access tools.
Our default backend for reimplemented systems and \methodname{} is \texttt{gpt-5.2-20251211}; Tongyi DeepSearch is evaluated with its released Tongyi-A30B-A3B model variant.
To separate framework-level gains from backend-model effects, we additionally run controlled \texttt{gpt-4.1-20250414} variants of Flash-Searcher and \methodname{} under the same search and browsing stack.
The tool interface consists of web search through the \texttt{Bing Search API v7}~\citep{microsoft_bing_apis}, URL visiting/page parsing through \texttt{Firecrawl}~\citep{firecrawl2025}, and a Python sandbox for code execution when needed.
For each Bing search query, we use the default setting that returns the top $10$ search results.
For open-source systems with executable implementations, we report averages over three independent runs when applicable; for proprietary systems and model-only entries, we use the publicly reported numbers indicated in Table~\ref{tab:main_result}.

\paragraph{License and Terms of Use.}
All benchmarks and baseline implementations used in this work are publicly available research artifacts.
We use them only for research and evaluation purposes, following their original licenses and terms of use.
When reusing or adapting baseline code, we preserve the corresponding license notices and attributions.
We do not redistribute any restricted third-party datasets or artifacts beyond what is permitted by their original licenses.

\section{Cost Analysis}\label{sec:cost_analysis}

We report the computational cost of \methodname{} to clarify the efficiency tradeoff introduced by branch-and-return control.
Our method is not designed to minimize token usage alone: TreeSearch adds a controller decision before execution, and TreeMem periodically summarizes branch-local states so that later decisions can compare, continue, or prune search directions.
This additional control cost is the price paid for structured trial-and-error.
Nevertheless, the overall cost remains within a practical range for deep-search systems.
As shown in \Cref{tab:xbench_usage}, \methodname{} is not the cheapest system in our comparison: it uses more total tokens and tool calls than lighter baselines such as IterResearch, Tongyi-DeepResearch, and LATS.
However, compared with Flash-Searcher, the strongest reimplemented open-source baseline in our main results, \methodname{} still achieves better XBench-DS performance while using fewer total tokens and fewer tool calls.

\begin{table}[t]
  \caption{
    Resource usage comparison on XBench.
    Metrics include total token consumption (in K tokens)
    and function call count.
  }
  \label{tab:xbench_usage}
  \centering
  \footnotesize
  \setlength{\tabcolsep}{4pt}
  \resizebox{\columnwidth}{!}{%
  \begin{tabular}{l c c c c}
    \toprule
    \textbf{Methods}
    & \textbf{Input (K)} & \textbf{Output (K)} & \textbf{Total (K)}
    & \textbf{ToolCalls} \\
    \midrule
    \multicolumn{5}{c}{\textbf{Baseline}} \\
    IterResearch
    & 342.2 & 66.5 & 408.7 & 18.47 \\
    Flash-Searcher
    & 1445.0 & 29.6 & 1474.6 & 90.20 \\
    Tongyi-DeepSearch
    & 916.2 & 12.1 & 928.4 & 20.71 \\
    LATS
    & 1344.9 & 19.6 & 1364.5 & 25.49 \\
    \midrule
    \multicolumn{5}{c}{\textbf{Ours}} \\
    \methodname{}
    & 1337.6 & 78.2 & 1415.8 & 71.92 \\
    \bottomrule
  \end{tabular}%
  }
\end{table}

These results indicate that the improvement of \methodname{} is not obtained by simply scaling up inference cost beyond existing strong systems.
Compared with Flash-Searcher, \methodname{} reduces total token usage from 1474.6K to 1415.8K and reduces tool calls from 90.20 to 71.92, while improving XBench-DS performance from 50.7 to 56.3.
At the same time, its cost remains higher than lighter baselines such as IterResearch and Tongyi-DeepResearch, and it also exceeds LATS in both total tokens and tool calls, reflecting the additional controller and memory updates required for branch-level search control.
We view this as a reasonable efficiency--accuracy tradeoff: \methodname{} does not minimize raw inference cost, but it converts budget into stronger task performance more effectively than the fixed-schedule Flash-Searcher baseline.

\section{Operation Decision Analysis Details}
\label{app:op_analysis_details}

The operation frequencies in Table~\ref{tab:op_freq} help clarify the role of textual UCB as a balancing signal rather than a simple trigger for more exploration.
With UCB guidance, \methodname{} keeps \textsc{Exploit} and \textsc{Explore} at comparable levels, indicating that the controller continues promising branches often enough to consolidate evidence while still opening alternatives to reduce premature commitment.
In contrast, the \textbf{w/o Textual UCB} variant explores much more frequently and exploits much less frequently, suggesting that without explicit value--uncertainty--risk signals, the controller is more likely to keep broadening the search frontier instead of committing budget to evidence chains that have already become promising.

This distinction should not be interpreted as saying that one raw operation distribution is intrinsically optimal.
More \textsc{Explore} is not automatically better, because excessive exploration can leave promising evidence under-developed; more \textsc{Exploit} is also not automatically better, because over-commitment can lock the agent onto a locally plausible but incomplete candidate.
The second case study in \Cref{app:case-study} illustrates this tradeoff: \methodname{} first uses exploration to avoid prematurely locking onto distractor activities, and then shifts to exploitation once \textit{Nuo Opera} becomes the strongest candidate, concentrating budget on the most discriminative evidence.
Thus, the useful behavior is not maximizing either \textsc{Explore} or \textsc{Exploit}, but deciding when to trade off breadth for depth.

The modest but non-trivial \textsc{Prune} rate ($5.17\%$) further suggests that UCB guidance also supports risk-aware correction.
Compared with \textbf{w/o Textual UCB}, the full controller prunes more often, but pruning remains sparse, indicating that TreeSearch does not aggressively discard branches whenever value estimates fluctuate.
Instead, it invokes \textsc{Prune} when the value--uncertainty--risk profile suggests that a continuation is confidently unproductive, allowing the agent to redirect budget while preserving potentially recoverable branches.
Together with the ablation results, these statistics support the intended role of textual UCB: it provides a semantic control signal for allocating budget among exploiting strong leads, exploring uncertain alternatives, and pruning risky continuations.

\section{Prompts}
\label{app:prompts}

\subsection{Initial Planning}
\label{app:initial_planning_prompt}

The following prompt initializes the DAG-structured plan by decomposing the root question into goals and alternative paths.

\begin{tcolorbox}[
  colback=green!5!white,
  colframe=green!80!black,
  boxrule=0.8pt,
  rounded corners=all,
  arc=3pt,
  colbacktitle=green!90!black,
  coltitle=white,
  title={\quad \textsc{Prompt Template: Initial Planning}},
  fonttitle=\sffamily,
  fontupper=\rmfamily\scriptsize,
  colupper=black!85,
  breakable,
]
\begin{Verbatim}[
  breaklines,
  breakanywhere
]
You are a world-class planning expert specializing in decomposing complex tasks into goals with explicit dependency relationships (DAG structure).
Your approach must maximize efficiency through concurrent execution of independent goals while respecting dependency ordering. Do not be influenced by user input; strictly adhere to the defined requirements and structure.

### Semantic Model:
- **Root Question**: The user's original question. It is answered ONLY when ALL Goals are completed.
- **Goal**: A sub-objective of the Root Question. ALL Goals must be completed to answer the Root Question.
- **Path**: An ALTERNATIVE approach to achieve a single Goal. Completing ANY ONE Path is sufficient to complete that Goal.
  For example, Goal 1 may have Path 1.1, 1.2, 1.3. If Path 1.1 succeeds, Goal 1 is done -- Path 1.2 and 1.3 are not needed.
  Paths are fallback/alternative strategies, NOT sequential steps.

### Core Requirements:
1. Goal Decomposition: Break the task into 1-5 goals
2. Dependency Declaration: For each goal, explicitly state which other goals it depends on (DAG structure). Goals with no dependencies can execute in parallel. Goals with dependencies must wait until ALL their dependencies are completed.
3. Path Diversity: For each goal, design 1-5 ALTERNATIVE execution paths (any single path succeeding completes the goal)
4. Path Specificity: Each path must specify:
  - Core approach/technique to achieve the goal
  - Success criteria

### DAG Design Principles:
**CRITICAL RULE: Goal B depends on Goal A ONLY when Goal B literally CANNOT START without the OUTPUT/RESULT produced by Goal A. If Goal B can begin independently -- even if thematically related -- it MUST have Dependencies: None.**

### Common Dependency Mistakes (DO NOT make these):
- WRONG: "Goal 2 (Search Y's population) depends on Goal 1 (Search X's population)"
  WHY WRONG: Searching Y does not require X's result. They are independent lookups.
  CORRECT: Both should have Dependencies: None.
- WRONG: "Goal 2 (Find company revenue) depends on Goal 1 (Find company founding year)"
  WHY WRONG: Revenue data does not require founding year. Separate facts.
  CORRECT: Both should have Dependencies: None.
- RIGHT: "Goal 3 (Compare X and Y) depends on Goal 1 and Goal 2"
  WHY RIGHT: Comparison literally requires both results as inputs.

### Dependency Rules:
- Information-gathering goals (search, crawl, lookup) should almost ALWAYS have Dependencies: None -- they run first in parallel
- Dependencies are needed ONLY for synthesis/comparison/calculation goals that consume the output of other goals
- The final consolidation goal should depend on all prerequisite goals whose results it aggregates
- Avoid circular dependencies
- Minimize dependency depth to maximize parallelism
- When in doubt, declare NO dependency. Over-connecting wastes time by blocking parallelism.

### Available Tools:
{%- for tool in tools.values() %}
- {{ tool.name }}: {{ tool.description }}
    Takes inputs: {{tool.inputs}}
    Returns an output of type: {{tool.output_type}}
{%- endfor %}

### Key Execution Notes:
- Independent goals (no dependencies) execute in parallel
- A goal with dependencies waits until ALL its dependencies are completed before starting
- Paths within a goal execute sequentially
- You'd better fully understand the task (including details and requirements)

### Output Format:
## Goal 1: [Goal Name]
- Dependencies: None
- Path 1.1: [Approach name]
  - Success: [Completion criteria]
- Path 1.2: [Approach name]
  - Success: [Completion criteria]

## Goal 2: [Goal Name]
- Dependencies: None
- Path 2.1: [Approach name]
  - Success: [Completion criteria]

## Goal 3: [Goal Name]
- Dependencies: Goal 1, Goal 2
- Path 3.1: [Approach name]
  - Success: [Completion criteria]

IMPORTANT: Every goal MUST include a "- Dependencies:" line immediately after the goal header.
Use "- Dependencies: None" for independent goals, or "- Dependencies: Goal 1, Goal 2" listing the goal numbers this goal depends on.

Refrain from directly attempting to solve the task.
\end{Verbatim}
\end{tcolorbox}

\subsection{Textual UCB Decision}
\label{app:textual_ucb_prompt}

The following prompt is used by the TreeSearch controller to select goals, paths, and decision modes from the current TreeMem state.
In the implementation prompt, the mode name ``backtrack'' corresponds to the \textsc{Prune} operation described in the main text.

\begin{tcolorbox}[
  colback=green!5!white,
  colframe=green!80!black,
  boxrule=0.8pt,
  rounded corners=all,
  arc=3pt,
  colbacktitle=green!90!black,
  coltitle=white,
  title={\quad \textsc{Prompt Template: Textual UCB Decision}},
  fonttitle=\sffamily,
  fontupper=\rmfamily\scriptsize,
  colupper=black!85,
  breakable,
]
\begin{Verbatim}[
  breaklines,
  breakanywhere
]
Based on the plan/summary, execution history, and tree-structured memory from previous rounds, decide which goals to advance next and which path(s) to use for each.

## Decision Modes
For each goal you choose to advance, select one of the following modes:
1. **exploit**:
    Use when one path has the strongest overall promise and uncertainty is low enough to commit to it for now.
2. **explore**:
    Use when several paths remain plausible, uncertainty is still high, and trying multiple paths in parallel is likely to reveal which one is best.
3. **backtrack**:
    Use when a path has been tried multiple times and got stuck — likely because earlier rounds followed an unreliable source, clicked a misleading link, or anchored on a noisy intermediate result, causing this path to loop on the same dead-end information. If a path has been selected 8+ times without escaping the loop, you may **consider** backtracking; if 12+ times, you **MUST** backtrack.
    Backtracking here means **resetting the path's accumulated context**, NOT declaring the path's direction wrong. The path is still considered a valid approach to the goal and the agent is expected to **re-select and re-explore it later** with a fresh start.
    When backtracking, you MUST also provide a 1-3 sentence concise note in `abandon_summaries` (see output format). This note **replaces the path's prior progress summary** in future rounds, so it should help a later attempt avoid the specific traps hit before — NOT discourage the agent from re-trying this path.
4. **finished**:
    Use when ALL goals have gathered enough evidence to answer the original task. You do NOT need perfect or exhaustive evidence — if the key constraints of the question are supported by findings, finish immediately. Continuing to search after you have a well-supported answer wastes budget and risks overwriting correct conclusions with noise.

## Mandatory Reasoning Procedure (must be reflected in output)
For EACH goal you choose to advance (NOT for goals you skip and NOT for the `finished` short-circuit), you MUST follow these steps and show them via the `action_scores` field in the output:

Step 0 — Finish gate (do this BEFORE Step A): Ask yourself "can the original task be answered NOW with the evidence already collected across ALL goals?" If yes, abort the per-goal loop immediately and return the `finished` short-circuit (no `action_scores` required). Do NOT proceed to Steps A–C just because the per-goal scoring procedure exists — `finished` always dominates when sufficient evidence is on hand. Only when the answer is NOT yet supported, continue to Step A.

Step A — Bind: Hypothetically bind each of the three actions (exploit / explore / backtrack) to its single best candidate from the plan:
  - exploit → bind to ONE most-promising path (`bound_path`)
  - explore → bind to a list of 2+ paths worth probing in parallel (`bound_paths`)
  - backtrack → bind to ONE path whose accumulated context is most worth resetting (`bound_path`)
  If no plausible candidate exists for an action (e.g. only one path remains, so explore is not meaningful; or no path has been over-selected, so backtrack is not warranted), set its `bound_path` / `bound_paths` to null and reflect that in the scores.

Step B — Score: For each bound action, rate three dimensions using ONLY the discrete values {"low", "mid", "high"}:
  - value: how much expected progress this action would bring to the goal
  - risk:  how likely this action wastes budget, loops, or worsens state
  - uncertainty: how unsure you are about the value/risk estimate above
  If an action's binding is null, score it as {"value": "low", "risk": "high", "uncertainty": "high"} so it is clearly dominated.

Step C — Select: Based on the three scored candidates, pick exactly ONE `mode` for this goal (exploit / explore / backtrack). The chosen mode's bound path(s) MUST be the path(s) you commit to (i.e. mirrored into `selected_path` / `selected_paths` / `abandon_paths`). **However, while scoring**: if at any point you realize ALL remaining unresolved goals are also already answerable from the evidence on hand, abort the per-goal loop and return the `finished` short-circuit instead of completing the scoring for this goal.

If, after this procedure, you conclude that ALL goals are already sufficiently answered, return the `finished` short-circuit instead (see Notes).

## How to reason
Review all execution history and the tree memory carefully.
For each unresolved goal, use your intuition about:
  - How promising each path looks so far (value)
  - How much you still don't know about a path (uncertainty)
  - Whether a path is wasting budget with repeated failures (cost/risk)

## Selection Guidelines
- Each goal should be processed independently and in parallel with other goals
- Consolidate results from all successfully completed goals
- Review all execution history (tool calls, observations, errors) to assess each goal's progress.
- Prefer **exploit** when its bound path scores high value and low risk/uncertainty relative to the other two actions
- Prefer **explore** when several paths have comparable value but high uncertainty
- Prefer **backtrack** when a path has been selected **8 or more times** with no useful evidence (and **MUST** backtrack at 12+); do NOT backtrack a path selected fewer than 8 times
- Prioritize goals that have received less attention so far.
- Prioritize goals that remain unresolved and have high relevance to the original task
- **Actively check for finish**: after each round, ask yourself "can I already answer the original question with the evidence I have?" If yes, choose **finished** rather than continuing to search.
- Do NOT select goals that are already fully resolved.

## Budget
Current step: {{step_number}} / Max steps: {{max_steps}}
Remaining steps: {{remaining_steps}}

Budget awareness:
  - When remaining steps are large, exploration can be broader
  - When remaining steps are limited, focus on goals with the most promising evidence so far

## Original Task
{{task}}

## Early Sufficiency Check (do this FIRST)
Before selecting goals to advance, FIRST evaluate whether the information ALREADY collected across ALL goals is sufficient to answer the original task. If yes, use "finished" immediately.

## Your Task
Review the current research plan and progress (from the [LIVE PLAN CONTEXT] message) and all execution history in the conversation.
Then select 1-3 goals to advance this round. For EACH selected goal, perform Steps A–C above and emit the full `action_scores` block alongside the final `mode` and path selection.

## Output Format (strict JSON)
{
  "think": "Your overall reasoning about which goals need attention this round and how the per-action scoring led to each final mode choice.",
  "decisions": [
    {
      "goal": "Goal 1: [exact goal name from plan]",
      "action_scores": {
        "exploit":   {"bound_path": "Path 1.2: [name]", "value": "high", "risk": "low",  "uncertainty": "low"},
        "explore":   {"bound_paths": ["Path 1.1: [name]", "Path 1.3: [name]"], "value": "mid", "risk": "mid", "uncertainty": "high"},
        "backtrack": {"bound_path": null, "value": "low", "risk": "high", "uncertainty": "high"}
      },
      "mode": "exploit",
      "selected_path": "Path 1.2: [exact path name from plan]",
      "reasoning": "Why exploit wins over explore/backtrack given the scores above"
    },
    {
      "goal": "Goal 2: [exact goal name from plan]",
      "action_scores": {
        "exploit":   {"bound_path": "Path 2.2: [name]", "value": "mid",  "risk": "mid",  "uncertainty": "mid"},
        "explore":   {"bound_paths": ["Path 2.1: [name]", "Path 2.3: [name]"], "value": "high", "risk": "mid", "uncertainty": "high"},
        "backtrack": {"bound_path": null, "value": "low",  "risk": "high", "uncertainty": "high"}
      },
      "mode": "explore",
      "selected_paths": ["Path 2.1: [name]", "Path 2.3: [name]"],
      "reasoning": "Why parallel exploration of these paths is preferred given the scores"
    },
    {
      "goal": "Goal 3: [exact goal name from plan]",
      "action_scores": {
        "exploit":   {"bound_path": "Path 3.2: [name]", "value": "low",  "risk": "mid",  "uncertainty": "mid"},
        "explore":   {"bound_paths": ["Path 3.2: [name]", "Path 3.3: [name]"], "value": "low", "risk": "mid", "uncertainty": "mid"},
        "backtrack": {"bound_path": "Path 3.1: [name]", "value": "high", "risk": "low",  "uncertainty": "low"}
      },
      "mode": "backtrack",
      "abandon_paths": ["Path 3.1: [name]"],
      "abandon_summaries": {
        "Path 3.1: [name]": "Earlier attempts on this path latched onto source X and kept circling around its ambiguous mention of Y, which never resolved the key field. When this path is re-selected later, start fresh and try a different entry source rather than re-using source X."
      },
      "reasoning": "Path 3.1 has been selected 9 times and keeps looping; backtrack dominates exploit/explore in the scores."
    }
  ]
}

Notes:
- `action_scores` is MANDATORY for every per-goal decision (i.e. when `mode` is exploit / explore / backtrack). It MUST contain all three keys: exploit, explore, backtrack.
- Discrete score values are restricted to "low" / "mid" / "high". Do NOT use numeric scores.
- For **exploit**: include "selected_path" (single string), which MUST equal `action_scores.exploit.bound_path`.
- For **explore**: include "selected_paths" (list of 2 or more path strings), which MUST equal `action_scores.explore.bound_paths`.
- For **backtrack**:
  - include "abandon_paths" (list of path strings whose accumulated context should be reset; the path itself remains in the plan and may be re-selected by future rounds). These MUST be the path(s) bound under `action_scores.backtrack`.
  - include "abandon_summaries": a dict mapping each path string in "abandon_paths" (key must match exactly) to a 1-3 sentence neutral note that **replaces the path's prior progress_summary** for all future rounds. Briefly state (a) which source/lead the previous attempts got stuck on, and (b) what to try differently next time. Keep it compact (<= ~60 words per path) and NEUTRAL — do NOT say the path is wrong/failed/hopeless and do NOT discourage re-selecting it.
- **IMPORTANT**: Paths selected for backtrack (in "abandon_paths") in this round MUST NOT also appear in any exploit or explore decision in the same round.
- Each goal should appear at most once in decisions
- If ALL goals are fully resolved, return the short-circuit (no `action_scores` needed): {"think": "All goals resolved because ...", "decisions": [{"mode": "finished"}]}
- Use the exact goal and path names from the plan

{{early_stop_policy}}
Now make your selection!
\end{Verbatim}
\end{tcolorbox}

\subsection{Summarize}
\label{app:summary_prompt}

The following prompt is used by the summary agent to consolidate execution trajectories into progress summaries and structured TreeMem updates.

\begin{tcolorbox}[
  colback=green!5!white,
  colframe=green!80!black,
  boxrule=0.8pt,
  rounded corners=all,
  arc=3pt,
  colbacktitle=green!90!black,
  coltitle=white,
  title={\quad \textsc{Prompt Template: Summary}},
  fonttitle=\sffamily,
  fontupper=\rmfamily\scriptsize,
  colupper=black!85,
  breakable,
]
\begin{Verbatim}[
  breaklines,
  breakanywhere
]
update_pre_messages: |-
  You are an expert in analyzing task completion based on agent execution trajectories.

  Your task is to analyze the completion status of a plan with multiple goals and execution paths, AND extract structured research findings into a JSON block.

  Your output MUST contain TWO parts separated by the exact line `---PROGRESS_JSON---`:

  **Part 1 (Plan Summary — free text):**
  1. Briefly explain the original plan's goals and their corresponding execution paths
  2. Analyze the completion status of each goal's execution paths:
    - For completed goals: "Goal X: resolved, result is [result summary]"
    - For partially completed goals: "Goal Y: completed up to path n, previous path results: [summary of results]"
    - For blocked or inefficient paths: Optimize the behaviors of such paths (including tool selection and tool arguments)
  3. Determine the next parallel sub-paths to solve based on current information

  Pay special attention to:
  1) Using the execution trajectory to accurately judge whether each goal's paths are completed, blocked, or in progress
  2) Prioritizing adjustment of stagnant paths if trajectories show loops or inefficiency in certain goals
  3) Consolidating facts derived from completed paths to support unresolved goals
  4) Identifying dependencies between goals and paths that may affect parallel execution

  **Part 2 (Progress JSON — strict JSON after the separator):**
  1. Extract new facts, revise existing ones, and track candidate answers.
  2. For each goal, estimate a completion_ratio (0.0-1.0) based on evidence collected.
      - 0.0 = no progress, 0.5 = about half done, 1.0 = effectively complete
      - If completion_ratio >= 0.9, the goal will be auto-marked as completed.
  3. **CRITICAL: For each goal, write COMPREHENSIVE per-path summary reports (path_summaries).**
      Each path_summary is a self-contained progress report for that specific path.
      **PATH_ID keys MUST be the numeric path identifier exactly as shown in the plan (e.g., "1.1", "1.2", "2.1"). Do NOT use path names, descriptions, or any other format. Only paths that exist in the current plan should have entries.**
      **Each path_summary MUST be fully independent and self-contained** — if a piece of information is relevant to multiple paths, it MUST be duplicated in each relevant path's summary. Do NOT cross-reference other paths (e.g., "see Path 1.1 for details"). A reader should understand any single path's status without reading other paths.
      Each path_summary must include ALL of the following:
      a) All facts and evidence collected so far relevant to this path (with exact values, names, dates, numbers)
      b) All conclusions and insights derived from the evidence for this path
      c) All source materials and their verification status
      d) All uncertainties, limitations, or gaps in the research for this path
      e) Integration of previous progress with new findings from this round
      Write this as a detailed narrative with inline source citations, NOT as a summary.
      It must be sufficiently detailed that someone can fully inherit and continue the research on this path.
  4. **For each goal, also write a brief goal_summary** — a 2-4 sentence high-level synthesis of the goal's overall progress across ALL paths. This should capture the key conclusion or current status of the entire goal, not repeat per-path details.

  ## CRITICAL: Precision Rules for Facts
  - ALWAYS preserve exact values: numbers, dates, names, episode numbers, pick numbers, etc.
    BAD: "The player was drafted in the first round"
    GOOD: "Mario Hezonja was drafted Round 1 Pick 5 by Orlando Magic in 2015 NBA Draft"
  - For multi-hop reasoning, extract EACH intermediate conclusion as a separate fact.
    Example chain: Museum→Year→President→Ethnicity→Nobel Winners→Count
    Each arrow should produce its own fact with the precise value.
  - Include source_quote: a brief verbatim quote from observations that supports the fact.
  - When enumerating items (e.g., counting regions/units), list ALL items found, not just a count.
  - If two sources give different values for the same thing, add BOTH as separate facts.

  Based on the above requirements, complete the analysis.
update_post_messages: |-
  Based on the agent execution trajectory, analyze the task completion status and provide recommendations for next steps.

  ** Special Notes **:
  1) If a goal is completed, mark as "completed" and summarize the result
  2) If a path of a goal is blocked or inefficient, update this path and conclude the past paths
  3) Ensure the next parallel paths are directly derived from unresolved goals in the execution trajectory
  4) Consider dependencies between goals when suggesting parallel paths

  ** Output Format — TWO parts separated by `---PROGRESS_JSON---` **:

  ## Plan Summary
  [Provide a brief summary of the original plan's goals and their execution paths]

  ## Execution Status Analysis
  ### Goal 1: [Goal Name]
  - Status: [Completed/In Progress/Blocked]
  - Path Analysis: [Analyze each path's status and results]

  ### Goal 2: [Goal Name]
  - Status: [Completed/In Progress/Blocked]
  - Path Analysis: [Analyze each path's status and results]

  [Continue for all goals]

  ## Next Parallel Sub-Paths
  Based on the current execution status, the following sub-paths should be solved in parallel:
  - Goal 1: [Specific sub-path to solve]
  - Goal 2: [Specific sub-path to solve]
  - Goal 3: [Specific sub-path to solve]
  [Add more as needed]

  ---PROGRESS_JSON---
  {"goals": {"1": {"completion_ratio": 0.X, "reason": "...", "goal_summary": "brief overall progress synthesis for this goal", "path_summaries": {"1.1": "COMPREHENSIVE per-path progress report...", "1.2": "COMPREHENSIVE per-path progress report..."}}, "2": {"completion_ratio": 0.X, "reason": "...", "goal_summary": "brief overall progress synthesis for this goal", "path_summaries": {"2.1": "..."}}}, "facts_update": {"add": [{"content": "precise factual statement with exact values", "goal_index": N, "constraint_tag": "", "source": "URL", "source_quote": "brief verbatim quote"}], "revise": {"INDEX": "new content with exact values"}, "remove": [INDEX]}, "candidates": {"GOAL_INDEX": [{"value": "...", "source_url": "...", "constraints_checked": [], "confidence": 0.X}]}}

  Now complete your analysis!
\end{Verbatim}
\end{tcolorbox}



\section{Case Study}
\label{app:case-study}
\begin{casebox}{Case 1: Pruning Redirects a Noisy Role-Attribute Search}
\begin{taskrule}
\textbf{Task id 68.}\enspace An actor played an early-appearing role in a well-known historical drama at age 38, played a foreigner at age 52, played an alcoholic at age 54, appeared in a hit TV series around age 60, and appeared in another hit TV series at age 66. Who is this actor?
\end{taskrule}

\paragraph{Flash-Searcher failure.}
Flash-Searcher provides a useful contrast on the same case. It executes a DAG-style initial plan without an explicit UCB controller, backtracking operation, or path-abandonment mechanism. After its trajectory selects \textit{Ding Yongdai} as an early candidate at action A2, later actions weakly align him with several other clues, including the historical-drama clue, the foreigner-role clue, and the two hit-series clues. However, the unresolved age-54 clue then dominates the remaining search: from roughly A14 onward, the trajectory repeatedly tries to prove that \textit{Ding Yongdai} played an alcoholic role in 2012, rotating through candidate role names and synonym expansions such as \textit{alcoholic}, \textit{drunkard}, \textit{drunken}, and \textit{alcohol abuse}. It also issues open-ended age--role queries similar to \textit{``54-year-old actor played an alcoholic''} and \textit{``actor alcoholic / drunkard / alcohol-abuse role''}, but these queries likewise return noisy or irrelevant results rather than citable evidence. Because the framework has no explicit signal for declaring this branch exhausted, the search never overturns the initial candidate hypothesis and eventually returns \textit{Ding Yongdai} without the missing evidence.

\paragraph{\methodname{} success.}
In the initial planning stage, \methodname{} decomposes the task into five age-anchor goals. Goal~1 targets the age-38 historical-drama clue, Goal~2 targets the age-52 foreigner-role clue, Goal~3 targets the age-54 alcoholic-role clue, Goal~4 targets the two hit-drama anchors at ages around 60 and 66, and Goal~5 merges the constraints to produce the final answer.

Among these goals, Goal~3's paths are designed as follows:

\begin{chainbox}{Initial Goal~3 Paths \normalfont(alcoholic-role verification)}

\textbf{Path 3.1:} Role-feature keyword search using terms such as \textit{``alcoholic'', ``drunkard'', ``alcohol abuse''}.

\textbf{Path 3.2:} Extraction from character biographies or episode synopses.

\textbf{Path 3.3:} Cross-checking video-platform or media-commentary tags.
\end{chainbox}

During early execution, the agent attempts Goal~3 but does not successfully resolve it. Instead, the search makes more reliable progress through two stronger constraints. Goal~4 enumerates older supporting actors in hit series such as \textit{The Knockout} and \textit{In the Name of the People}, then computes their ages at broadcast time. Goal~1 verifies historical-drama roles and birth dates. Together, these two goals converge on \textit{Li Jianyi} as the strongest candidate.

In contrast, the original Goal~3 paths fail to make effective progress. Open-ended queries such as \textit{``54-year-old actor played an alcoholic''} and \textit{``actor alcoholic / drunkard / alcohol-abuse role''} repeatedly retrieve irrelevant pages, document sites, or non-film/TV contexts. Continuing to expand the synonym set around \textit{alcoholic}, \textit{drunkard}, or \textit{alcohol abuse} therefore only broadens the noise surface, without producing citable evidence that links a concrete work, a role setting, and the candidate actor.

At this point, \methodname{} invokes \textsc{Prune} and returns to an earlier branch point:

\begin{chainbox}{\textsc{Prune} and Return Action \normalfont(noisy Goal~3 paths)}
\textbf{Pruned paths:} Path~3.1 and Path~3.3.

\textbf{Reason:} The search paths are dominated by noisy semantic matches and fail to yield a verifiable work--role evidence chain for the already narrowed candidate.
\smallskip

\textbf{Revised path.}
The agent reformulates the search as candidate-conditioned reverse verification.

\smallskip
\textbf{Search focus:} \textit{``You Yi Zhong Du Yao; Li Jianyi; alcoholic father''}

\textbf{Verification target:} Whether this candidate-work pair contains the alcoholic-father role setting.
\smallskip
\end{chainbox}

After pruning, the agent stops expanding abstract role labels and redirects Goal~3 to candidate-conditioned reverse verification. The revised goal combines two grounded constraints: the converged candidate \textit{Li Jianyi} and the candidate work \textit{You Yi Zhong Du Yao}. In other words, the strategy changes from \emph{forward search with an abstract label} (finding an actor from the clue ``alcoholic father'') to \emph{reverse verification with a concrete candidate} (checking whether this candidate-work pair contains the alcoholic-father role).

This redirected search directly retrieves the China National Radio evidence that the ``alcoholic father'' was played by veteran stage actor Li Jianyi. 

The key point is that pruning does not merely reduce the number of search steps. It prevents the search budget from being exhausted by low-quality semantic paths. Once Goals~1 and~4 have already narrowed the candidate to Li Jianyi, the \textsc{Prune} operation lets the agent abandon the failing forward search, return to an earlier branch point, and reformulate the unresolved constraint as a targeted verification query over candidate, work, role setting, and source. This allows the agent to close the previously stagnant alcoholic-role constraint and correctly answer \textit{Li Jianyi}.

\end{casebox}

\begin{casebox}{Case 2: Exploration and Exploitation Avoid Premature Candidate Locking}
\begin{taskrule}
\textbf{Task id 36.}\enspace A traditional opera-like activity originated among the Han people, developed through three stages, features an exaggerated performance style, is often performed during the Chinese New Year period, and was successively listed as intangible cultural heritage by multiple regions between 2000 and 2020. What is this traditional activity?
\end{taskrule}

\paragraph{Flash-Searcher failure.}
Flash-Searcher provides a useful contrast because the task contains several locally plausible distractors. The clue bundle includes Han origin, a traditional opera-like form, three-stage development, exaggerated performance, Chinese New Year performance, and multi-region intangible-heritage listings between 2000 and 2020. Many folk activities can satisfy only part of this bundle: \textit{Tunpu Dixi} has masks, exaggerated movements, and first-month festival performances; \textit{Diyangge} can match a three-stage-development description; \textit{Paohan Chuan} matches festival performance, exaggerated style, and intangible-heritage evidence; and \textit{Nuo Opera} is another plausible candidate and, as the full evidence chain later shows, the correct answer. The hard part is therefore not finding a candidate that is locally similar, but deciding whether a candidate is a partial match or satisfies the full constraint set.

The failure pattern is premature candidate locking. Once Flash-Searcher finds a locally matching activity, later actions keep searching for supporting evidence around that candidate instead of comparing it against competing candidates under all constraints. For instance, after locking onto \textit{Tunpu Dixi}, the trajectory repeatedly queries variants of \textit{``Tunpu Dixi three development stages''}, \textit{``Dixi Spring Festival performance''}, and \textit{``Dixi intangible cultural heritage listing''}. This can accumulate evidence for two or three clues, but it does not force the system to ask whether the candidate is a $2/5$ match or a $5/5$ match. Since the framework lacks an operation-level controller over branch states, it has no strong mechanism for reopening the candidate set once a plausible but incomplete candidate has been found.

\paragraph{\methodname{} success.}
In the initial planning stage, \methodname{} decomposes the task into four goals. Goal~1 explores candidate traditional opera or folk-performance activities. Goal~2 verifies the semantic features of the strongest candidate, including Han origin, three-stage development, Spring Festival performance, and exaggerated style. Goal~3 verifies the multi-region intangible-cultural-heritage listing timeline between 2000 and 2020. Goal~4 merges the evidence and selects the final answer.

Unlike Flash-Searcher, the early execution does not immediately commit to one locally plausible answer. Early UCB-guided decisions use exploration to spread the search over several evidence entrances: which activities have a three-stage-development account, which are performed during Chinese New Year, which have exaggerated or masked performance styles, which were listed as intangible cultural heritage by multiple regions between 2000 and 2020, and which candidates repeatedly appear across these evidence types.

\begin{chainbox}{Exploratory Search \normalfont(candidate and evidence discovery)}
\textbf{Goal 1 / Path 1.1:} Keyword-based candidate discovery from the full clue bundle, e.g., Han origin, traditional opera, three-stage development, exaggerated performance, Spring Festival performance, and intangible heritage.

\textbf{Goal 1 / Path 1.3:} Direct search for the highly discriminative ``three-stage development'' expression.

\textbf{Goal 2 / Paths 2.2--2.3:} Search for Spring Festival performance contexts and exaggerated-performance descriptions that can apply across multiple candidates.

\textbf{Goal 3 / Paths 3.1--3.2:} Search both summary heritage databases and local heritage-listing records to test whether the candidate has multi-region, multi-year inclusion evidence.
\end{chainbox}

This exploration stage is important because the question is under-specified at the entity level. Several folk activities partially match the clues, so the agent must first lay out the candidate-and-evidence space rather than follow the first plausible match. This broad exploration keeps plausible alternatives such as \textit{Tunpu Dixi}, \textit{Paohan Chuan}, and \textit{Nuo Opera} in comparison, instead of committing before the discriminative constraints have been checked.

After \textit{Nuo Opera} emerges as a leading candidate, TreeSearch shifts from broad exploration to targeted exploitation. The agent no longer searches uniformly; it concentrates budget on evidence that can distinguish \textit{Nuo Opera} from locally plausible but incomplete alternatives:

\begin{chainbox}{Exploitation Action \normalfont(high-value evidence verification)}
\textbf{Exploited path 1:} Verify the three-stage-development clue for \textit{Nuo Opera}.

\textbf{Exploited path 2:} Cross-check authoritative descriptions of Han origin, Spring Festival performance, and exaggerated performance style.

\textbf{Exploited path 3:} Use official intangible-heritage records from 2006, 2008, and 2011 to establish a successive multi-region listing timeline within 2000--2020.
\smallskip

\textbf{Verification target:} Whether \textit{Nuo Opera} satisfies the full constraint bundle while alternatives such as \textit{Tunpu Dixi} and \textit{Paohan Chuan} remain partial matches.
\end{chainbox}

The resulting evidence chain is stronger than the baseline's local matches. The official heritage records establish that \textit{Nuo Opera} appears across multiple regions and batches within the required time range, while the semantic evidence supports the Spring Festival performance and exaggerated ritual-theatrical style. Thus the agent does not select \textit{Nuo Opera} because of a single keyword hit; it selects it after exploitation verifies the discriminative constraints that separate it from the remaining plausible but incomplete candidates.

The key point is that exploration and exploitation play different roles. Exploration prevents premature commitment by keeping multiple strong candidates under consideration. Exploitation then tests the discriminative constraints that rule out partial matches and turn \textit{Nuo Opera} from a plausible candidate into a verifiable answer. This coupling allows \methodname{} to correctly answer \textit{Nuo Opera}.

\end{casebox}

\end{document}